\documentclass[%
11pt,
onecolumn,preprintnumbers,pre, superscriptaddress
preprint,
nofootinbib,
amsmath,amssymb,
]{revtex4-1}

\usepackage[english]{babel}
\usepackage[utf8]{inputenc}
\usepackage[babel]{csquotes}
\usepackage[T1]{fontenc}
\usepackage{amsthm}
\usepackage{amsmath}
\usepackage{amsfonts}
\usepackage{mathrsfs}
\usepackage{amssymb}
\usepackage{mathtools}
\usepackage{graphicx}
\usepackage{xcolor}
\usepackage{subfigure}
\usepackage{geometry}
\usepackage{braket}
\usepackage{sidecap}
\usepackage{float}
\usepackage{bm}
\usepackage{lipsum}
\usepackage{bbold}
\usepackage{url}

\theoremstyle{remark}

\begin{document}
\title{Machine learning in spectral domain.}

\author{Lorenzo Giambagli$^1$, Lorenzo Buffoni$^{1,2}$, \\
Timoteo Carletti$^3$, Walter Nocentini$^1$, \\
Duccio Fanelli$^1$}

\affiliation{$1.$ Universit\`{a} degli Studi di Firenze, Dipartimento di Fisica e Astronomia,
CSDC and INFN, via G. Sansone 1, 50019 Sesto Fiorentino, Italy}
\affiliation{$2.$ Dipartimento di Ingegneria dell'Informazione, Universit\`{a} di Firenze,
Via S. Marta 3, 50139 Florence, Italy}
\affiliation{$3.$ naXys, Namur Institute for Complex Systems, University of Namur, 8 Rempart de la Vierge, B5000 Namur, Belgium}

%\pacs{} 

\begin{abstract}
Deep neural networks are usually trained in the space of the nodes,  by adjusting the weights of existing links via suitable optimization protocols. We here propose a radically new approach 
which anchors the learning process to reciprocal space. Specifically, the training acts on the spectral domain and seeks to modify the {eigenvalues and eigenvectors} of transfer operators in 
direct space. The proposed method is ductile and can be tailored to return either linear or non-linear classifiers. {Adjusting the eigenvalues, when freezing the eigenvectors entries, yields performances which are superior to those attained with standard methods {\it restricted} to a operate with an identical number of free parameters. Tuning the eigenvalues correspond in fact to performing a global training of the neural network, a procedure which promotes (resp. inhibits) collective modes on which an effective information processing relies.  This is at variance with the usual approach to learning which implements instead a local modulation of the weights associated to pairwise links. Interestingly, spectral learning limited to the eigenvalues returns a distribution of the predicted weights which is close to that obtained when training the neural network in direct space, with no restrictions on the parameters to be tuned. Based on the above, it is surmised that spectral learning bound to the eigenvalues could be also employed for pre-training of deep neural networks, in conjunction with conventional machine-learning schemes.  Further, linear processing units inserted within adjacent, non-linearly activated, layers produce an effective enlargement of the set of trainable eigenvalues. After training, the added layers can be retracted, as a sort of telescopic booms, returning compact networks with improved classification performances. In our current implementation, and to recover a feed-forward architecture in direct space, we have postulated a nested indentation of the eigenvectors. This choice allows one to visualise the successive embedding of the processed information, from the input to the output, as mutually entangled gears of an algorithmic device which is operated in dual space. Changing the eigenvectors to a different non-orthogonal basis alters the topology of the network in direct space and thus allows to export the spectral learning strategy to other frameworks, as e.g. reservoir computing.}
\end{abstract}

\maketitle

\section{Introduction}
Machine learning (ML) \cite{bishop_pattern_2011,cover_elements_1991,hastie2009elements,hundred} refers to a broad field of study, with multifaceted applications of  cross-disciplinary  breadth. ML {is a subset of Artificial Intelligence (AI) which}  ultimately aims at developing computer algorithms that improve automatically through experience. The core idea is that systems can learn from data, so as to identify distinctive patterns and make consequently decisions, with minimal human intervention. The range of applications of ML methodologies is extremely vast \cite{sutton2018reinforcement,graves2013speech,sebe2005machine,grigorescu2020survey}, and still growing at a steady pace due to the pressing need to cope with the 
efficiently handling of big data \cite{chen2014big}. Biomimetic approaches to sub-symbolic AI \cite{rosenblatt1961principles} inspired the design of powerful algorithms. These latter sought to reproduce the unconscious process underlying fast perception, the neurological paths for rapid decision making, as e.g. employed for faces \cite{meyers2008using} or spoken words \cite{caponetti2011biologically} recognition. 

An early example of a sub-symbolic brain inspired AI was the perceptron \cite{rosenblatt1958perceptron}, the influential ancestor of deep neural networks (NN) \cite{bengio2007greedy,Goodfellow-et-al-2016}. The perceptron is indeed an algorithm for supervised learning of binary classifiers. It is a linear classifier, meaning that its forecasts are based on a  linear prediction function which combines a set of weights with the feature vector. Analogous to neurons, the perceptron adds up its input: if the resulting sum is above a given threshold the perceptron fires (returns the output the value 1) otherwise it does not (and the output equals zero). Modern multilayer perceptrons, account for multiple hidden layers with  non-linear activation functions. The learning is achieved via {minimizing the classification error.} 
{Single or multilayered perceptrons should be trained by examples \cite{bengio2007greedy,hinton2006fast,rumelhart1988learning}.  Supervised learning requires indeed a large set of positive and negative examples, the training set, labelled with their reference category.} 

The perceptrons' acquired ability to perform classification is eventually stored in a finite collection of numbers, the weights and thresholds that were learned during the successive epochs of the supervised training.  To date, it is not clear how such a huge collection of numbers (hundred-millions of weights in state of the art ML applications) are synergistically interlaced for the deep networks  
to execute the assigned tasks, with an exceptional degree of robustness and accuracy \cite{xie2020explainable,hinton2015distilling,erhan2010understanding}. 

Starting from these premises, the aims of this  paper are multifold. On the one side, we will develop a novel learning scheme which is anchored on reciprocal space. Instead of {iteratively} adjusting the weights of the edges that define the connection among nodes, we will modify {the spectra of a collection of suitably engineered matrices that bridge adjacent layers. To eventually recover a multilayered feedforward architecture in direct space, we postulate a nested indentation of the associated eigenvectors. These latter act as the effective gears of a processing device operated in reciprocal space. The directed indentation between stacks of adjacent eigenvectors yield a compression of the activation pattern, which is eventually delivered to the detection nodes.} 

{As a starting point, assume eigenvectors are frozen to a reference setting which fulfills the prescribed conditions. The  learning is hence solely restricted to the eigenvalues, a choice which amounts to performing a {\it global} training, targeted to identifying key collective modes, the selected eigen-directions,  for carrying out the assigned classification task. The idea of conducting a global training on a subset of parameters has been also proposed in other works \cite{frankle2020training, Gabri__2019}. This is at odd with the usual approach to machine learning where {\it local} adjustments of pairwise weights are implemented in direct space.   As we shall prove, by tuning the  eigenvalues, while freezing the eigenvectors, yields performances superior to those reached with usual (local) techniques bound to operate with an identical number of free parameters, within an equivalent  network architecture. Eigenvalues are therefore identified as key target of the learning process, proving more fundamental than any other set of  identical cardinality, allocated in direct space. Remarkably, the distribution of weights obtained when applying the spectral learning technique restricted to the eigenvalues is close to that recovered when training the neural network in direct space, with no restrictions on the parameters to be adjusted. In this respect, spectral learning bound to the eigenvalues could provide a viable strategy for pre-training of deep neural networks. Further,  the set of trainable eigenvalues can be expanded at will by inserting linear processing units between the adjacent layers of a non-linear multilayered perceptron. Added linear layers act as veritable booms of a {\it telescopic neural network}, which can be extracted during the learning phase and retracted in operational mode, yielding compact networks with improved classification skills. The effect of the linear expansion is instead negligible, if applied to neural learning of standard conception. The entries of the indented eigenvectors can be also trained resulting in enhanced performance, as compared to the setting where eigenvalues are exclusively modulated by the learning algorithm. To demonstrate the principles which underly  spectral training, we employ the MNIST database, a collection of handwritten digits to be classified. The examined problem is relatively simple:  a modest number of tunable parameters is indeed necessary for achieving remarkable success rates. When allowing for the simultaneous training of the eigenvalues and (a limited fraction of ) eigenvectors, the neural networks quickly saturates to accuracy scores which are indistinguishable from those obtained via conventional approaches to supervised learning. More challenging tasks should be probably faced to fully appreciate the role played by a 
 progressive optimization of the eigenmodes, the collective directions in reciprocal space where information flows.  As remarked above, the eigenvectors have been here constructed so as to yield a feedforward multi-layered architecture in direct space. By relaxing this assumption, comes to altering the network topology and thus exporting the spectral learning strategy to other frameworks, as e.g. reservoir computing. In general terms,  working in the spectral domain corresponds to optimizing a set of {\it non orthogonal} directions (in the high dimensional space of the nodes) and associated weights (the eigenvalues), a global outlook which could contribute to shed novel light on the theoretical foundations of supervised learning.}

{\section{Linear and non-linear spectral learning}}

\begin{figure}
\centering
\includegraphics[scale=0.3]{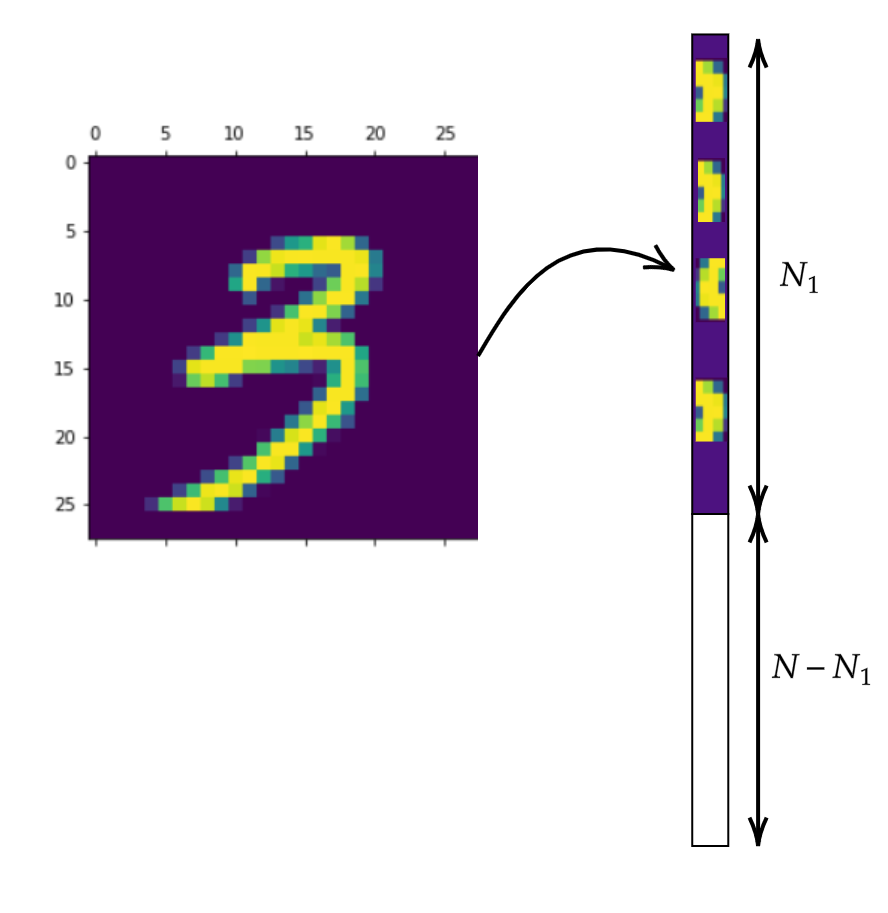}  
\caption{\it  Each image of the training set is mapped into a column vector $\vec{n}_1$, of size $N$, whose first $N_1 = 28 \times$ 28 entries are the intensities displayed on the pixels of the image.}
\label{fig1} 
\end{figure}

To introduce and test the proposed method we will consider a special task, i.e. recognition of handwritten digits. To this end, we will make use of the MNIST database \cite{lecun1998mnist} which has a training set of 60,000 examples, and a test set of 10,000 examples. Each image is made of $N_1=28 \times 28$ pixels and each pixel bears an 8-bit numerical intensity value, see Fig. \ref{fig1}. A deep neural 
network can be trained using standard backpropagation \cite{bengio2007greedy} algorithms to assign the weights that link the nodes (or perceptrons) belonging to consecutive layers. The first layer has $N_1$ 
nodes and the input is set to the corresponding pixel's intensity. The highest error rate reported on the original website of the database \cite{lecun1998mnist} is 12 \%, which is achieved using a simple linear classifier, with no preprocessing. In early 2020, researchers announced 0.16 \% error \cite{byerly2020branching} with a deep neural network made of branching and merging convolutional networks. Our goal here is to contribute to the analysis with a radically different approach to the learning, {rather than joining the efforts to break current limit in terms of performance and classification accuracy. More specifically, and referring to the MNIST database as a benchmark application, we will assemble a network made of $N$ nodes, organized in successive $\ell$ layers, tying the training to reciprocal space.} 

{Directed connections between nodes belonging to consecutive layers are encoded in a set of $\ell-1$, $N \times N$ adjacency matrices. The eigenvectors of these latter matrices are engineered so as to favour the information transfer from the reading frame to the output layer, upon proper encoding. The associated eigenvalues represent the primary target of the novel learning scheme. In the following we will set up the method, both with reference to its linear and non-linear versions. Tests performed on the MNIST database are discussed in the next Section.}

{\subsection{Linear spectral learning: Single-layer perceptron trained in reciprocal space}}  

Assume $N_i$ to label the nodes assigned to layer $i$, {and define} $N=\sum_{i=1}^{\ell} N_i$. {For the specific case here inspected the} output layer is composed by ten nodes ($N_{\ell}=10$), where recognition takes eventually place. Select one image from the training set and be $n_1$ $(=0,1,2..,9)$ the generic number therein displayed. We then construct a column vector $\vec{n}_1$, of size $N$, whose first $N_1$ entries are the intensities displayed on the pixels of the selected image (from the top-left to the bottom-right, moving horizontally), as illustrated in Fig. \ref{fig1}. {All other entries are initially set to zero. As we shall explain in the following, our goal is to transform the input $\vec{n}_1$ into an output vector with same dimensions. The last  $N_{\ell}$ elements of this latter vector represent the output nodes where reading is eventually performed.} 

To set the stage, we begin by reporting on a simplified scenario that, as we shall prove in the following, yields a single layer perceptron. The extension to multi-layered architectures will be discussed {right after}.

\begin{figure}
\centering
\includegraphics[scale=0.5]{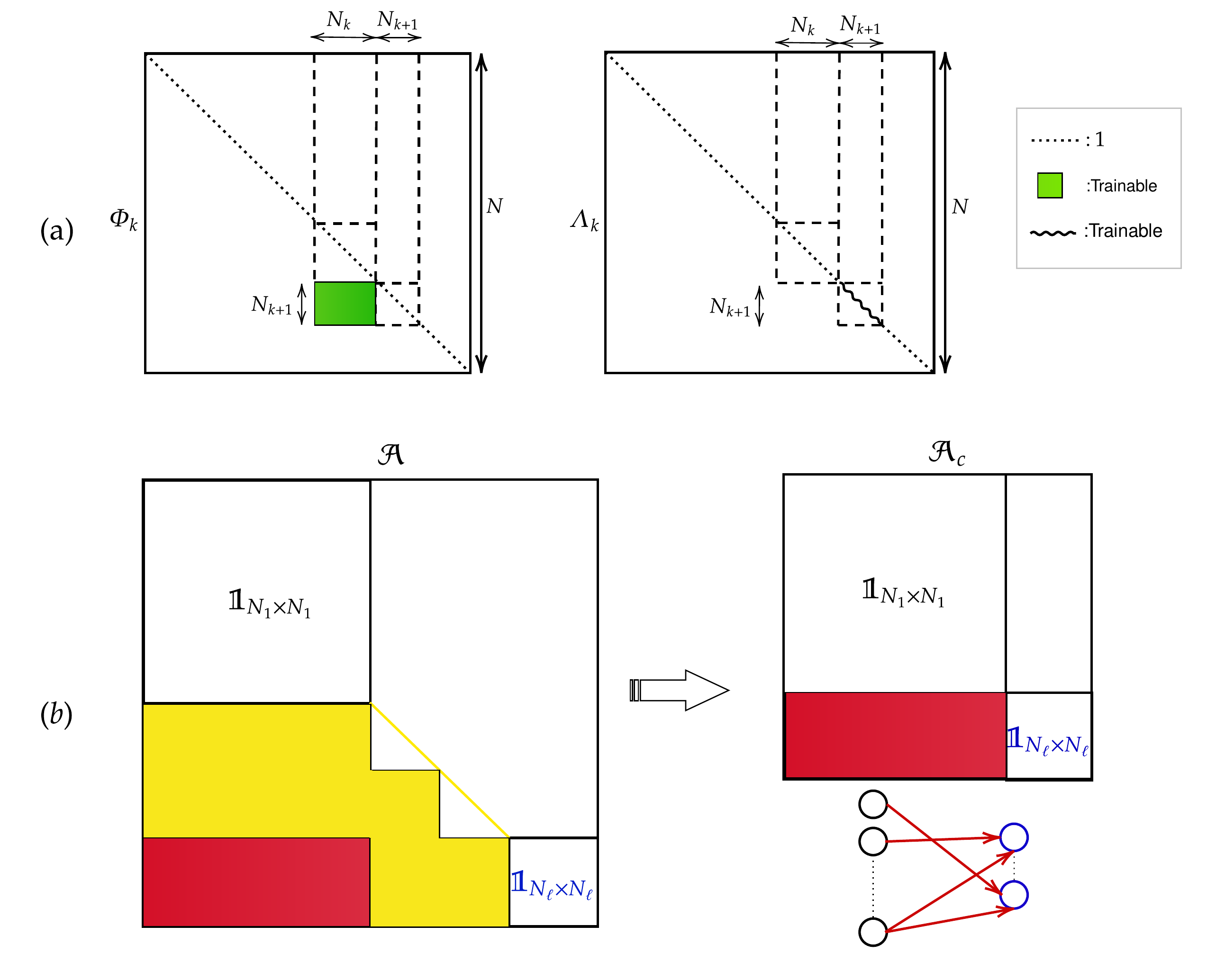} 
\caption{\it  {Panel (a): the structure of matrix $\Phi_k$ is schematically depicted. The diagonal entries of $\Phi_k$ are unities. The sub-diagonal block of size 
 $N_{k+1} \times N_{k}$ for $k=1,\ell-1$ is filled with uniform random numbers in $[a,b]$, with $a,b \in \mathbb{R}$. These blocks yields an effective indentation between successive stacks of linearly independent eigenvectors. The diagonal matrix of the eigenvalues $\Lambda_k$ is also represented. The sub-portions of $\Phi_k$ and $\Lambda_k$ that get modified by the training performed in spectral domain are  highlighted (see legend). In the experiments reported in this paper the initial eigenvectors entries are uniform random variables distributed in $[-0.5,0.5]$. The eigenvalues are uniform random numbers distributed in the interval $[-0.01,0.01]$. Optimizing the range to which the initial guesses belong (for both eigenvalues and eigenvectors) is an open problem that we have not tackled.
 Panel (b): a $(N_1 + N_\ell) \times (N_1 + N_\ell)$ matrix $\mathcal{A}_c$ can be obtained from $\mathcal{A}=\left( \Pi_{k=1}^{\ell-1} \mathbf{A}_{k}  \right) $, which provides the weights for a single layer perceptron, that maps the input into the output, in direct space.}}
\label{fig2} 
\end{figure}

{Consider the entry layer made of $N_1$ nodes and the outer one composed of $N_2$ elements. In this case $N=N_1+N_2$. The input vector $\vec{n}_1$ undergoes a linear transformation to yield $\vec{n}_2=\mathbf{A}_1 \vec{n}_1$ where $\mathbf{A}_1$ is a $N\times N$ matrix that we shall characterize in the following. Introduce matrix $\Phi_1$: this is the identity matrix $\mathbb{1}_{N \times N}$ modified by the inclusion of a sub-diagonal block $N_{2} \times N_{1}$, e.g. filled with uniformly distributed random numbers, defined in a bounded interval, see Fig. \ref{fig2}. The columns of $\Phi_1$, hereafter $\left(\vec{\phi}_1\right)_k$ with $k=1,...,N$, define a basis of the $N$ dimensional space to which $\vec{n}_1$ and $\vec{n}_2$ belong.  Then, we introduce the diagonal matrix  $\Lambda_1$. The entries of $\Lambda_1$ are set to random (uniform) numbers spanning a suitable interval. A straightforward calculation returns $\left(\Phi_1\right)^{-1}=2 \mathbb{1}_{N \times N}-\Phi_1$. We hence define  $\mathbf{A}_1= \Phi_1 \Lambda_1 \left(2 \mathbb{1}_{N \times N}-\Phi_1\right)$ as the matrix that transforms $\vec{n}_1$ into $\vec{n}_2$.  Because of the specific structure of the input vector, and owing the nature of $\mathbf{A}_1$, the information stored in the first $N_1$ elements of $\vec{n}_1$ is passed to the $N_2$ successive entries of $\vec{n}_2$, in a compactified form which reflects both the imposed eigenvectors' indentation and the chosen non trivial eigenvalues. }

{To see this more clearly, expand the $N$-dimensional input vector $\vec{n}_1$ on the basis made of $\left( \vec{\phi}_1 \right)_k$ to yield $\vec{n}_1=\sum_{k=1}^N c_k \left( \vec{\phi}_1 \right)_k$ where $c_k$ stands for the coefficients of the expansion. The first  $N_1$ vectors are necessarily engaged to explain the non zero content of $\vec{n}_1$ and, because of the imposed indentation,  rebound on the successive $N_2$ elements of the basis. These latter need to adjust their associated weights $c_k$  to compensate for the echoed perturbation.  The action of  matrix $\mathbf{A}_1$ on the input vector $\vec{n}_1$ can be exemplified as follows:} 

\begin{equation}
\vec{n}_2 = \mathbf{A}_1 \vec{n}_1 = \mathbf{A}_1 \sum_{k=1}^N c_k \left( \vec{\phi}_1 \right)_k = \sum_{k=1}^{N_1+N_2} c_k \left( \Lambda_1 \right)_k \left( \vec{\phi}_1 \right)_k
\end{equation}

{where $\left( \Lambda_1 \right)_k$ are the element of matrix $\Lambda_1$. In short, the entries of $\vec{n}_2$ from position $N_1+1$ to position $N_1+N_2$ represent a compressed (if $N_2<N_1$) rendering of the supplied input signal, the key to decipher the folding of the message being stored in the $N_{2} \times N_{1}$ sub-diagonal block of $\Phi_1$, (i.e. the eigenvector indentation) and in the first set of $N=N_1+N_2$ eigenvalues $\left( \Lambda_1 \right)_k$. The key idea is to propagate this message passing scheme, from the input to the output in a multi-layer setting,  and adjust (a subset of) the spectral parameters involved so as to optimize the encoding of the information.}

{ To this end, we introduce the $N \times N$ matrix operator $\Phi_k$, for $k=2,...,\ell-1$. In analogy with the above, $\Phi_k$ is the identity matrix $\mathbb{1}_{N \times N}$ modified with a sub-diagonal block $N_{k+1} \times N_{k}$, which extends from rows $N_k$ to $N_k+N_{k+1}$, and touches tangentially the diagonal, as schematically illustrated in Fig. \ref{fig2} (a). Similarly, we introduce $\Lambda_k$, for $k=2,...,\ell-1$, which is obtained from the identity matrix $\mathbb{1}_{N \times N}$ upon mutating to uniformly distributed random entries the diagonal elements that range from $\sum_{i=1}^k N_i$ (not included) to $\sum_{i=1}^{k+1} N_i$ (included). Finally, we define $\mathbf{A}_k= \Phi_k \Lambda_k \left(2 \mathbb{1}_{N \times N}-\Phi_k\right)$, as the matrix that transforms $\vec{n}_k$ into $\vec{n}_{k+1}$, with $k=2,...,\ell-1$. In principle, both non trivial eigenvalues' and eigenvectors'  input can be self-consistently adjusted by the envisaged learning strategy. The input signal $\vec{n}_1$ is hence transformed into an output vector $\vec{n}_{\ell}$ following a cascade of linear transformations implemented via matrices $\mathbf{A}_k$. In formulae:}

\begin{equation}
\vec{n}_{\ell} =  \mathbf{A}_{\ell-1} ...\mathbf{A}_{1}  \vec{n}_1 = \left( \Pi_{k=1}^{\ell-1}  \Phi_k \Lambda_k \left(2 \mathbb{1}_{N \times N}-\Phi_k\right)\right) \vec{n}_1
\end{equation}

{where in the last step we made use of the representation of $\mathbf{A}_k$ in dual space. The generic vector $\vec{n}_{k+1}$, for $k=1,..., \ell-1$ is obtained by applying matrix $\mathbf{A}_k$  to  $\vec{n}_k$. The first 
$N_1+N_2+...+N_k$ components of $\vec{n}_{k+1}$ coincide with the corresponding entries of $\vec{n}_k$, namely $\left[ \vec{n}_{k+1} \right]_m \equiv \left[ \vec{n}_{k} \right]_m$ for $m<N_1+N_2+...+N_k$. Here, 
$\left[ \left(\vec{\cdot}\right) \right]_m$ identifies the $m$-th component of the vector $\left(\vec{\cdot}\right)$. Recall that, by construction, $\left[ \vec{n}_{k} \right]_m=0$ for $m> N_1+N_2+...+N_k$. On the contrary, the components 
$\left[ \vec{n}_{k+1} \right]_m$ with $N_1+N_2+...+N_k+1<m<N_1+N_2+...+N_k+N_{k+1}$ are populated by non trivial values which reflect the eigenvectors indentation, as well as the associated eigenvalues.  This observation can be mathematically proven as follows. Write $\vec{n}_k$ on the basis formed by the eigenvectors  
 $\left( \vec{\phi}_k \right)_l$ to eventually get:}
 
\begin{equation}
\vec{n}_k = \sum_{l=1}^{N_1+N_2+...+N_{k+1}} c_l \left( \vec{\phi}_k \right)_l \equiv \sum_{l=1}^{N_1+N_2+...+N_k} c_l \vec{e}_l 
\end{equation}
 
{where $\left(\vec{e}_1, \vec{e}_2...\right)$ stand for the canonical basis and the last inequality follows the specific structure of the eigenvectors (remark that the leftmost sum in the above equation includes   $N_{k+1}$ more elements than the second). By definition:}

\begin{equation}
\vec{n}_{k+1} = \mathbf{A}_k \vec{n}_k = \sum_{l=1}^{N_1+N_2..+N_{k+1}} c_l \left( \Lambda_k \right)_l \left( \vec{\phi}_k \right)_l
\end{equation}

{From the above relation, one gets for $m \le N_1+N_2+...+N_k$}

\begin{equation}
\left[ \vec{n}_{k+1} \right]_m  =  \sum_{l=1}^{N_1+N_2..+N_{k}} c_l \left[ \vec{e}_l \right]_m \equiv \left[ \vec{n}_{k} \right]_m
\end{equation}

{where the first equality sign follows from the observation that $\left( \vec{\phi}_k \right)_l$ coincides with  $\vec{e}_l$ and  $\left( \Lambda_k \right)_l=1$, 
over the explored range of $m$.  For  $N_1+N_2+...+N_k+1 \le m \le N_1+N_2+...+N_k+N_{k+1}$, we obtain instead:}

\begin{equation}
\left[\vec{n}_{k+1}\right]_m =  \sum_{l=N_1+N_2..+N_{k-1}}^{N_1+N_2..+N_{k+1}} c_l \left( \Lambda_k \right)_l \left[ \left( \vec{\phi}_k \right)_l \right]_m
\end{equation}

{Finally, it is immediate to show that $\left[\vec{n}_{k+1}\right]_m=0$ for $m> N_1+N_2+...+N_k+N_{k+1}$, because of the specific form of the employed eigenvectors. In short, the information contained in the last non trivial $N_k$ entries of $\vec{n}_k$ rebound on the successive $N_{k+1}$ elements of $\vec{n}_{k+1}$,  funnelling the information downstream from the input to the output.
The successive information processing relies on the indented (non orthogonal) eigenvectors and the associated eigenvalues,  which hence define the target of the training  in reciprocal space.
}

{ To carry out the learning procedure one needs to introduce a loss function  $L(\vec{n}_1)$. For illustrative purposes this latter can be written as:}

\begin{equation}
\label{LF}
L(\vec{n}_1) = \left\lVert l(\vec{n}_1) - \sigma\left[ \left( \Pi_{k=1}^{\ell}  \Phi_k \Lambda_k \left(2 \mathbb{1}_{N \times N}-\Phi_k\right)  \right) \vec{n} _1\right ] \right\lVert^2 
\end{equation}

{ where $\sigma(\cdot)$ is the softmax operation applied to the last entries of the $\ell$-th image of the input vector $\vec{n}_1$. In the above expression, $l(\vec{n}_1)$ stands for the label attached to $\vec{n}_1$ depending on its category. More into details, the $k$-th entry of $l(\vec{n}_1)$ is equal unit (and the rest identically equal to zero) if  the number supplied as an input is identical to $k$, with 
$k=0,1,...,9$. The loss function can be minimized by acting on the free parameters of the learning scheme. Specifically, the learning can be restricted to the set of $N$ non trivial eigenvalues, split in $\ell$ distinct groups, each referred to one of the $\mathbf{A}_{k}$
matrices (i.e. $N_1+N_2$ eigenvalues of $\mathbf{A}_{1}$, $N_3$ eigenvalues of $\mathbf{A}_{2}$,...., $N_{\ell}$ eigenvalues of $\mathbf{A}_{\ell-1}$). In addition, the 
sub-diagonal block entries of $\Phi_k$, the elements of the basis which dictate the successive indentation between adjacent layers, can be adjusted as follows the training scheme. In the following section we will report about the performance of the method, implemented in its different modalities, against those obtained with a classical approach to the learning anchored in direct space. In the actual implementation  we have chosen to deal with a categorical cross-entropy loss function.}

{Before ending this section a few remarks are mandatory. Introduce  $\mathcal{A} = \Pi_{k=1}^{\ell} \mathbf{A}_{k}$. The linear transformation that links the input vector $\vec{n}_1$ to the generated output $\vec{n}_{\ell}$, can be compactly expressed as $\vec{n}_{\ell} = \mathcal{A} \vec{n}_1$. Then, recall that the classification relies on examining the last $N_\ell$ entries of $\vec{n}_{\ell}$. 
Hence, for the specific setting here examined, where the mapping is obtained as a cascade of linear transformations, one can imagine to recast the whole procedure in a space of reduced dimensionality. Be 
$\vec{z}$ a column vector made of $N_1 + N_\ell$ elements.  The first $N_1$ entries of $\vec{z}$ are the intensities on the pixels of the selected image, as for the homologous $\vec{{n}}_1$ quantity. The other elements are set to zero. Then,  consider the $(N_1 + N_\ell) \times (N_1 + N_\ell)$ matrix $\mathcal{A}_c$ (the label $c$ stands for {\it compact}), constructed from $\mathcal{A}$ by trimming out all the information that pertain to the intermediate layers, as introduced in the reciprocal space (see Fig. \ref{fig2}(b)). Stated differently, matrix $\mathcal{A}_c$ provides the weighted links that feed from the input to the output layer in direct space, via the linear transformation $\mathcal{A}_c \vec{z}$: this is a single layer perceptron, shown in  Fig. \ref{fig2}(b), which was trained by endowing reciprocal space with an arbitrary number of additional dimensions, the intermediate stacks responsible for the sequential embedding of the information. Intermediate layers can be literally extracted, during the training phase, and subsequently retracted in operational mode. The importance to allowing for additional layers, and so provide the neural network of a telescopic attribute,  will be assessed in the forthcoming sections.}

{From the algorithmic point of view the process outlined above can be rephrased in simpler, although equivalent terms. For all practical purposes, one could take the (column) input vector $\vec{{n}}_1$ to have $N_1+N_2$ elements. Following the scheme depicted above, the first $N_1$ entries are the intensities on the pixels of the selected image, while the remaining $N_2$ elements are set to zero. We now introduce a $(N_1+N_2) \times (N_1+N_2)$  matrix $\mathbf{A}_{1}$. This is the identity matrix $\mathbb{1}_{(N_1+N_2) \times (N_1+N_2)}$ with the inclusion of a sub-diagonal block $N_{2} \times N_{1}$, which handles the information processing that will populate the second $N_2$ elements of the output vector $\vec{{n}}_2= \mathbf{A}_{1} \vec{{n}_1}$. Then, we formally replace the  $(N_1+N_2)$ column vector $\vec{{n}}_2$ with a column vector made of $(N_2+N_3)$ elements,  termed $\vec{{n}}_{2t}$, whose first $N_2$ elements are the final entries of $\vec{{n}}_2$. The remaining $N_3$ elements of $\vec{{n}}_{2t}$ are set to zero. Now, rename $\vec{{n}}_{2t}$ as $\vec{{n}}_2$ and presents it as the input of a $(N_2+N_3) \times (N_2+N_3)$  matrix $\mathbf{A}_{2}$, with a non trivial sub-diagonal $N_{3} \times N_{2}$ block. This latter maps the first $N_2$ elements of the input vector, into the successive $N_3$ of the output one, by completing the second step of an algorithmic scheme which can be iteratively repeated. In analogy with the above, each  $(N_k+N_{k+1}) \times (N_k+N_{k+1})$ matrix $\mathbf{A}_{k}$ can be written as $\mathbf{A}_k= \Phi_k \Lambda_k \left(2 \mathbb{1}_{(N_k+N_{k+1}) \times (N_k+N_{k+1})}-\Phi_k\right)$, where now the column vectors of $\Phi_k$ are the eigevenctors of  $\mathbf{A}_k$ and form a non-orthogonal basis of the $(N_k+N_{k+1})$
space where input and output vectors belong. $\Lambda_k$ is a diagonal matrix of the eigenvalues: the first $N_k$ are set to one, while the other $N_{k+1}$ are non trivial entries to be adjusted self-consistently via the learning scheme. Framing the process in the augmented space of $N$ dimensions, as done earlier, allows us to avoid adapting the dimensions of the involved vectors at each iteration. 
On the contrary, this is a convenient procedure to be followed when aiming at a numerical implementation of the envisaged scheme. Notice that to discuss the algorithmic variant of the method, we made use of the same symbols employed earlier. The notation clash is however solely confined to this paragraph.} 

{In the following, we will discuss how these ideas extend to the more general setting of non-linear multi-layered neural networks.}

 {\subsection{Training non-linear multi-layered neural networks in the spectral domain}}

{ In analogy with the above,  the image to be processes is again organized in a $N\times1$ column vector $\vec{n}_1$. This latter is transformed into $\vec{n}_2=\mathbf{A}_1 \vec{n}_1$, where matrix 
${N \times N}$ matrix $\mathbf{A}_1$ is recovered from its spectral properties, respectively encoded in $\Phi_1$ and $\Lambda_1$. The output vector $\vec{n}_2$ is now filtered via a suitable {\it non-linear} function $f(\cdot)$. This step marks a distinction between, respectively, the linear and non-linear versions of the learning schemes. For the applications here reported we have chosen to work with a rectified linear unit  (ReLU) $f(\cdot)= max(0, \cdot)$. Another possibility is to set $f(\cdot, \beta_1)=\tanh[\beta_1(\cdot)]$, where  $\beta_1$ is a control parameter which could be in principle self-consistently adjusted all along the learning procedure. We are now in a position to iterate the  same reasoning carried out in the preceding section, adapted to the case at hand. More specifically, we introduce the generic $N\times N$ matrix $\mathbf{A}_k= \Phi_k \Lambda_k \left(2 \mathbb{1}_{N \times N}-\Phi_k\right)$ which transforms $\vec{n}_k$ into $\vec{n}_{k+1}$, with $k=2,...,\ell-1$.  The outcome of this linear transformation goes through the non-linear filter. The loss function  $L(\vec{n})$ generalizes to:}

\begin{equation}
\label{LF}
L(\vec{n}) = \left\lVert l(\vec{n}_1) - \sigma\left( f\left(\mathbf{A}_{\ell-1}.... f\left (\mathbf{A}_2  f \left (\mathbf{A}_1 \vec{n_1},\beta_1 \right), \beta_2  \right),\beta_{\ell-1} \right) \right)	   \right\rVert^2
\end{equation}
 
\begin{figure}
\centering
\includegraphics[scale=0.5]{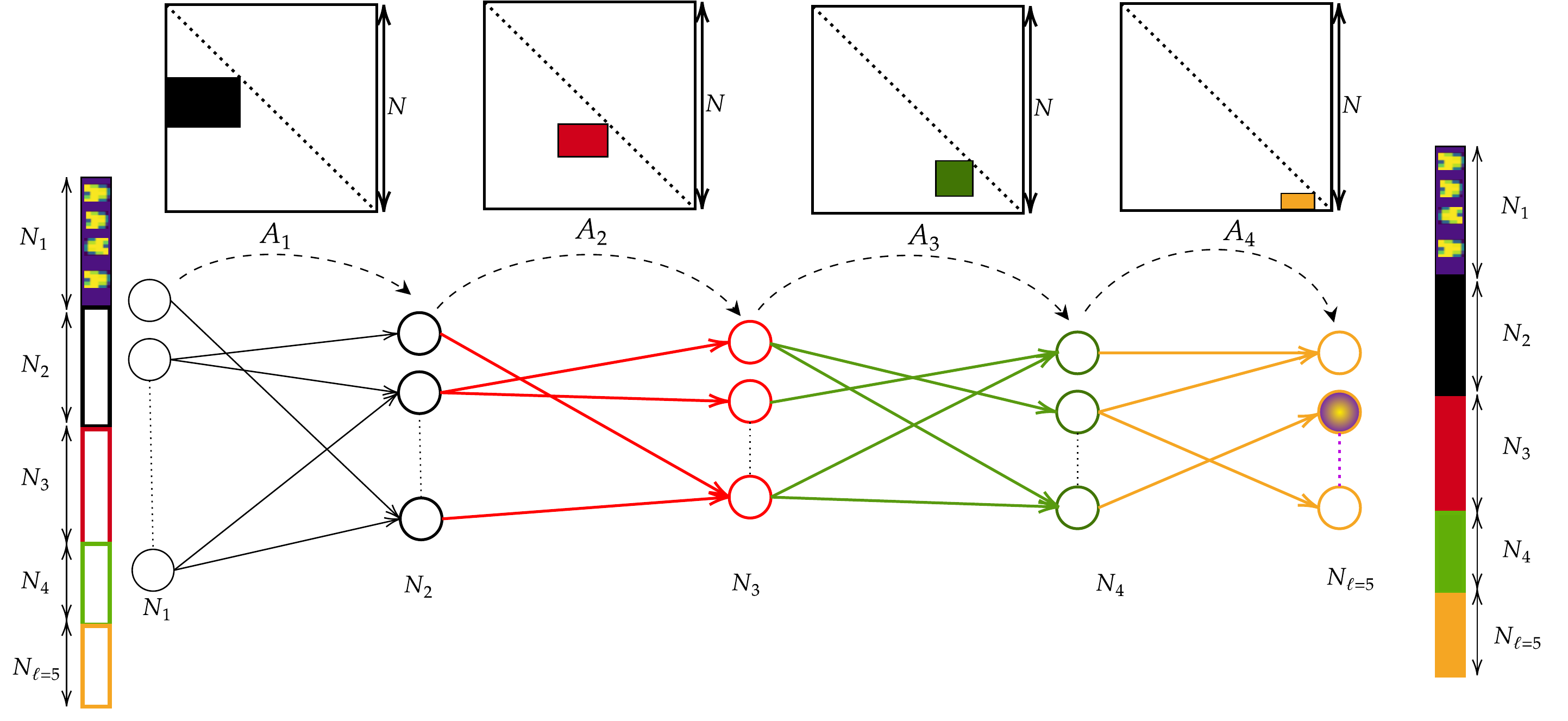} \\ 
\caption{\it  {The non-linear version of the training scheme returns a multi-layered architecture with self-loops links in direct space. Linear and non-linear transformation can be combined at will, matrices $\mathbf{A}_k$ providing the connection between successive layers. Linear layers can be retracted in operational mode, following a straightforward variant of the compactification procedure described in the main body of the paper. }}
\label{fig3} 
\end{figure}

{with an obvious meaning of the involved symbols. In the set of experiments reported below we assume, in analogy with the above, a categorical cross-entropy loss function. The  loss function is minimized upon adjusting  the free parameters of the learning scheme: the $\ell-1$ blocks of tunable eigenvalues, the elements that define the successive indentation of the nested basis which commands the transfer of the information  (and e.g. the quantities $\beta_k$, if the sigmoidal hyperbolic function is chosen as a non-linear filter). This eventually yields a fully trained network, in direct space, which can be unfolded into a layered architecture to perform pattern recognition (see Fig. \ref{fig3}). Remarkably, self-loop links are also present. The limit of a linear single layer perceptron is recovered when silencing the non-linearities:  a $(N_1 + N_\ell) \times (N_1 + N_\ell)$ matrix $\mathcal{A}_c$ can be generated from the $N \times N$ matrices $\mathbf{A}_{k}$, following the same strategy outlined above. A sequence of linear layers can be also interposed between two consecutive non-linear stacks. The interposed layers allow to enlarge the space of parameters employed in the learning scheme, and can be retracted when operating the deep neural network after completion of the learning stage. Their role is {\it de facto} encapsulated in the entries of the linear operator that bridges the gap between the adjacent non-linear stacks, as explained above when referring  to the telescopic operational modality.}

  {\section{Results}}

To build and train the aforementioned models we used TensorFlow and created a custom spectral layer matrix that could be integrated in virtually every TensorFlow or Keras model. That allowed us to leverage on the automatic differentiation capabilities and the built-in optimizers of TensorFlow. Recall that we aim at training {just a  a portion of the diagonal of $\Lambda_k$ and a block of $\Phi_k$}. To reach this goal we generated two fully trainable matrices, for each layer in the spectral domain, and applied a suitably designed mask to filter out the sub-parts of the matrices to be excluded from the training. This is easy to implement and, although improvable from the point of view of computational efficiency, it works perfectly, given the size of the problem to be handled. We then trained all our models with the AdaMax optimizer \cite{kingma2014adam} by using a learning rate of $0.03$ for the linear case and $0.01$ for the non-linear one. The training proceeded for {about $20$ epochs} and during each epoch the network was fed with batches of images of {different size, ranging from $300$ to $800$.}. These hyperparameters have been chosen so as to improve on GPU efficiency, accuracy and stability. However, we did not perform a systematic study to look for the optimal setting. All our models have been trained on a virtual machine hosted by Google Colaboratory. Standard neural networks have been trained on the same machine using identical software and hyperparameters, for a fair comparison. Further details about the implementation, as well as a notebook to reproduce our results, can be found in the public repository of this project \cite{gitrepo}.

\begin{figure}
\centering
\includegraphics[scale=0.7]{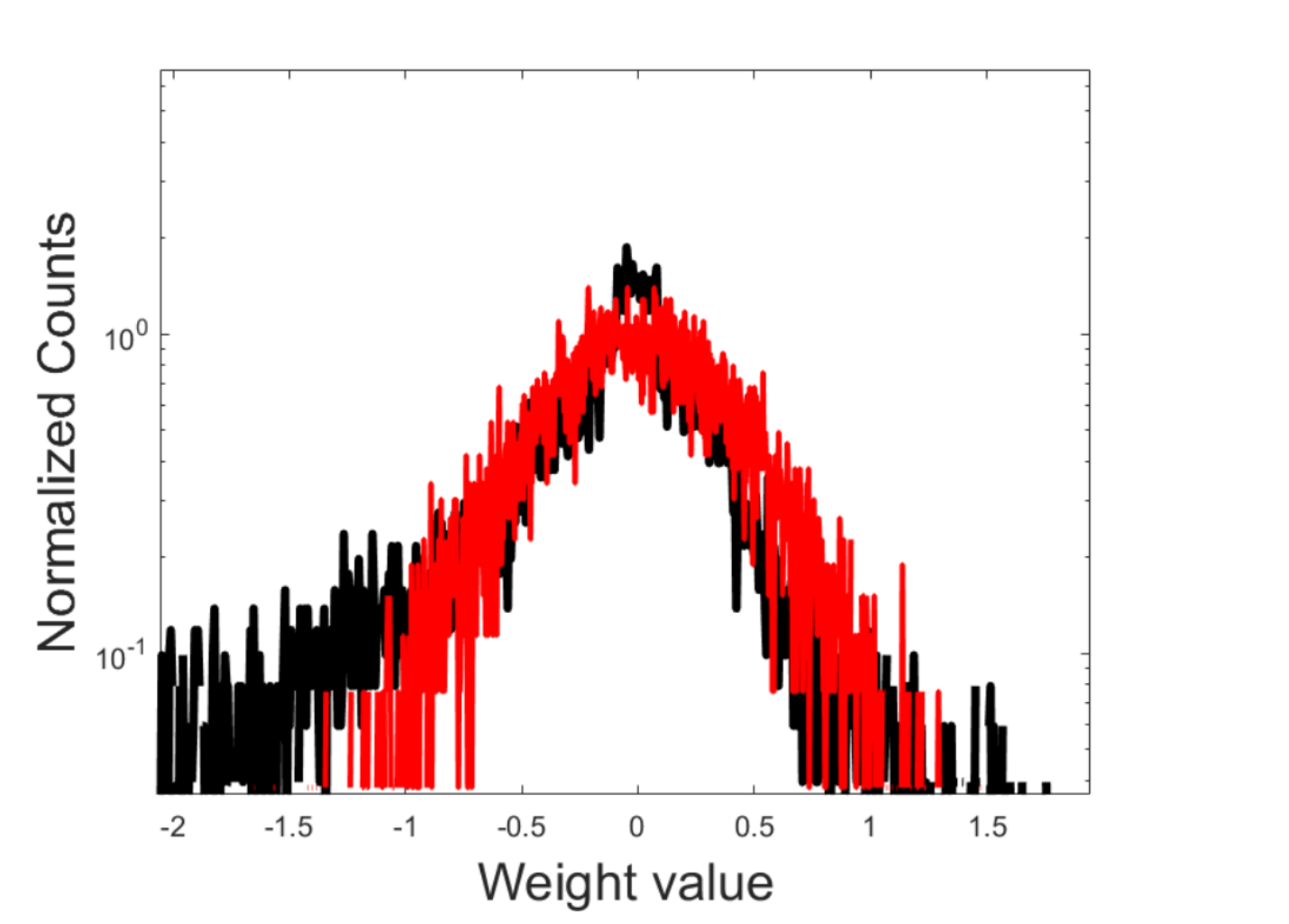} \\ 
\caption{\it  {Distribution of the weights of a perceptron. The red line follows the spectral training limited the $N_1+N_2$ eigenvalues. The black line follows the training in direct space. Here, $N_1 \times N_2$ parameters are adjusted in the space of the nodes. The distribution are very similar, but the spectral learning employs about $10 \%$ of the parameters used in direct space. The distributions obtained 
when forcing the training in direct space to operate on a subset of $N_1+N_2$ weights are very different from the one displayed (for every choice of the randomly selected family of weights to be trained).}}
\label{fig4} 
\end{figure}

{We shall start by reporting on the performance of the linear scheme. The simplest setting is that of a perceptron made of two layers: the input layer with $N_1=28 \times 28 = 784$ nodes and the output one made of $N_2=10$ elements. The perceptron can be trained in the spectral domain by e.g. tuning the $N=N_1+N_2=794$ eigenvalues of $\mathbf{A}_{1}$, the matrix that links the input ($\vec{n}_1$) and output ($\vec{n}_2$) vectors. The learning restricted to the eigenvalues returns a perceptron which performs the sought classification task with an accuracy (the fraction of correctly recognized images in the test-set)  of $(82 \pm 2) \%$ (averaging over $5$ independent runs). This figure is to be confronted with the accuracy of a perceptron trained with standard techniques in direct space. For a fair comparison, the number of adjustable weights should be limited to $N$. To this aim, we randomly select a subset of weights to be trained and carry out the optimization on these latter. The process is repeated a few ($5$ in this case) times and, for each realization, the associated accuracy computed. Combining the results yields an average performance of $(79 \pm 3) \%$ , i.e. a slightly smaller score (although compatible within error precision) than that achieved when the learning takes place in the spectral domain. When the training extends to all the $N_1 \times N_2$ weights (plus $N_1+N_2$ bias), conventional learning yields a final accuracy of $(92.7 \pm 0.1) \%$. This is practically identical to the score obtained in the spectral domain, specifically $(92.5 \pm 0.2) \%$, when the sub-diagonal entries of the eigenvectors matrix are also optimized (for a total of $N_1+N_2+N_1 \times N_2$ free parameters). The remarkable observation is however that the distribution of the weights as obtained when the learning is restricted on the eigenvalues (i.e using about the 10 \% of the parameters employed for a full training in direct space) matches quite closely that retrieved by means of conventional learning schemes, see Fig. \ref{fig4} . This is not the case when the learning in direct space acts on a subset of  $N$, randomly selected, weights (data not shown). Based on the above, it can be therefore surmised that optimizing the eigenvalues constitutes a rather effective pre-training strategy, which engages a modest computational load.}   

{To further elaborate on the potentiality of the proposed technique, we modify the simple two-layers  perceptron, with the inclusion of supplementary computing layers. As explained above the newly added layers plays an active role during the learning stage, but can be retracted in operating mode so as to return a two-layers perceptron. The weights of this latter bear however an imprint of the training carried out for the linear network in the expanded configuration. Two alternative strategies will be in particular contemplated. On the one side, we will consider a sole additional layer, endowed with $N_2$ nodes, interposed between the input and output layers made of, respectively, $N_1=784$ and $N_{\ell} \equiv N_3=10$ nodes. We will refer to this as to the {\it wide linear} configuration. The performance of the method can be tested by letting $N_2$ to progressively grow. On the other side,  the {\it deep linear} configuration is obtained when interposing a sequence of successive (linear) stacks between the input ($N_1=784$) and the output ($N_{\ell} =10$) layers.}

\begin{figure}
\centering
\includegraphics[scale=0.5]{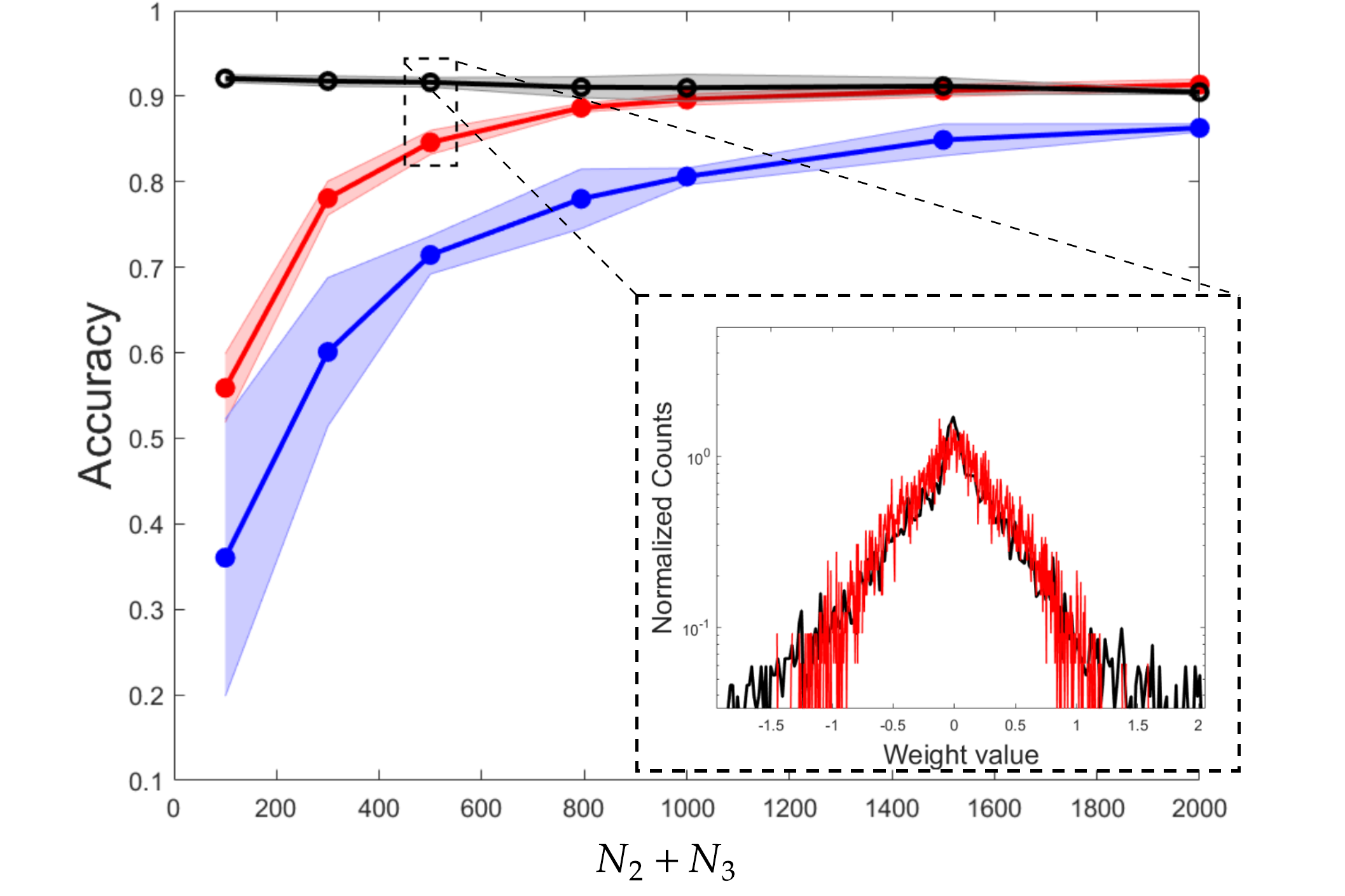} \\ 
\caption{\it  {A three layers neural network is considered. The accuracy of the neural network is plotted as a function of the number of parameters that we chose to train with the spectral algorithm, $N_2+N_3$. The red line reports on the performance of the spectral training. The blue line refers to the neural network trained in direct space: the optimization runs on $N_2+N_3$ parameters, a subset of the total number of adjustable weights $N_1 N_2+N_2 N_3$. The black line stands for the accuracy of the linear neural network when training the full set of $N_1 N_2+N_2 N_3$ parameters. Notice that the 
reported accuracy is comparable to that obtained for a standard two layers perceptron. Inset:
the distribution of the entries of  the equivalent perceptrons are plotted. The red curve refer to the spectral learning restricted to operate on the eigenvalues; the black profile to the neural network 
trained in direct space, employing the full set of adjustable parameters. In both cases, the weights refer to the two layers configuration obtained by retracting the intermediate linear layer employed during the learning stage. }}
\label{fig5} 
\end{figure}

 { In Fig. \ref{fig5},  we report on the performance of the wide learning scheme as a function  of $N_2+N_3$. As we shall clarify, this latter stands for the number of trained parameters for (i) the spectral learning acted on a subset of the tunable eigenvalues and for (ii) the conventional learning in direct space restricted to operate on a limited portion of the weights.  The red line in the main panel of Fig. \ref{fig5} refers to the simplified scheme where a subset of the eigenvalues are solely tuned (while leaving the eigenvectors fixed at the random realization set by the initial condition). We have in particular chosen to train the second bunch of $N_2$ eigenvalues of the transfer matrix $\mathbf{A}_{1}$ and the $N_3=10$ non trivial eigenvalues of matrix $\mathbf{A}_{2}$, in line with the prescriptions reported in the preceding Section. The blue line reports on the accuracy of the neural network trained in direct space: the target of the optimization is a subset of cardinality $N_2+N_3$ of the $N_1 N_2+N_2 N_3$ weights which could be in principle adjusted in the space of the nodes. The performance of the spectral method proves clearly superior, as it can be readily appreciated by visual inspection of Fig. \ref{fig5}. The black line displays the accuracy of the linear neural network when the optimization acts on the full set of $N_1 N_2+N_2 N_3$ trainable parameters. No improvement is detectable when increasing the size of the intermediate layer: the displayed accuracy is substantially identical to that obtained for the basic perceptron trained with $N_1 N_2=7840$ parameters. The spectral learning allows to reach comparable performance already at $N_2=1000$ ($13 \%$ of the parameters used for the standard two layers perceptron with $N_1 \times N_2$ parameters, as discussed above). 
% while at $N_2=2000$ $25 \%$ of the parameters used for the perceptron) the reported score appears slightly superior.
 In the inset of Fig. \ref{fig5}, the distribution of the entries of matrix $\mathcal{A}_c$, the equivalent perceptron, is depicted in red for the setting highlighted in the zoom. The black line refers to the two-layers equivalent of the neural network trained in direct space, employing the full set of trainable parameters (black dot enclosed in the top-left dashed rectangle drawn in the main panel of Fig. \ref{fig5}). The two distributions look remarkably close, despite the considerable reduction in terms of training parameters, as implemented in the spectral domain (for the case highlighted, $0.13 \%$ of the parameters employed under the standard training). Similarly to the above, the distribution obtained when forcing the training in direct space to act on a subset of $N_1+N_2$ weights are just a modest modulation of the initially assigned profile, owing to the {\it local} nature of the learning in the space of the nodes.} 
  
\begin{figure}
\centering
\includegraphics[scale=0.5]{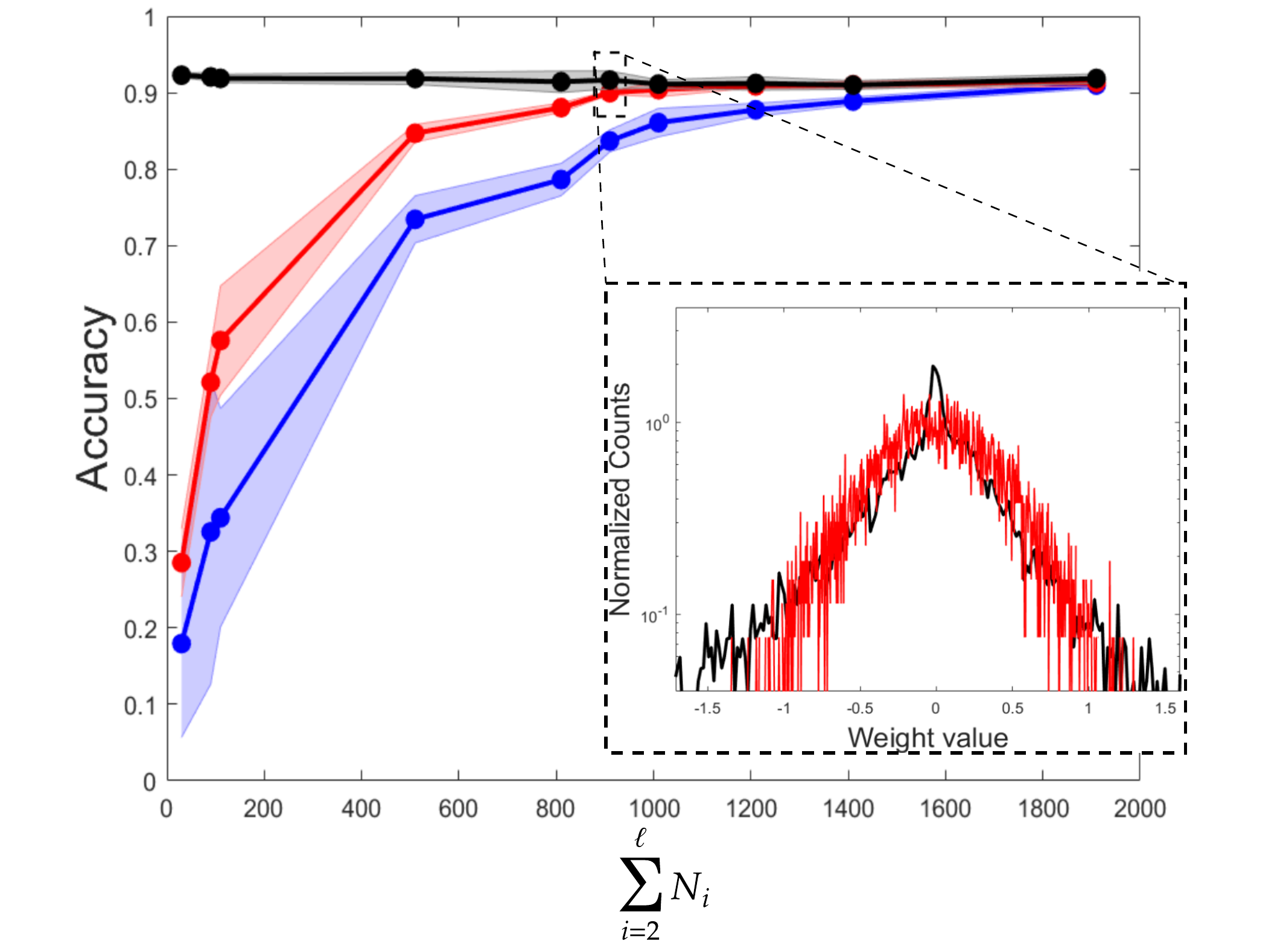} \\ 
\caption{\it  { The performance of the spectral algorithm are tested for a multi-layered linear configuration. Symbols are chosen in analogy to Fig. \ref{fig5}. In all cases, the input layer is made of $N_1=784$ elements and the output layer has $N_{\ell}=10$ nodes. The first five points, from left to right in each of the curves depicted in the main panel, refer to a three layers (linear) neural network. The size of the intermediate layer is progressively increased, as $N_2=20,80,100,500,800$. The total number of trained eigenvalues is $N_2+N_3$. The subsequent four points are obtained by considering a four layers architecture. In particular, $N_2=800$ while $N_3$ takes values in the interval ($100,200,400,600$). The training acts on $N_2+N_3+N_4$  eigenvalues.  The final point in each curve is obtained with a four layers deep neural network. Here, $N_2=800$, $N_3=600$ and $N_3=500$, for a total of $N_2+N_3+N_4+N_5$ tunable parameters in the spectral setting. Inset:
the distribution of the entries of the equivalent perceptrons are displayed, with the same color code adopted in Fig. \ref{fig5}. Also in this case,  the weights refer to the two layers configuration obtained by retracting the intermediate linear layers employed in the learning stage.
}}
\label{fig6} 
\end{figure}

 {In Fig. \ref{fig6}, we report the results of the tests performed when operating under the deep linear configuration. Symbols are analogous to those employed in Fig. \ref{fig5}. In all inspected cases, the entry layer is made of $N_1=784$ elements and the output one has $N_{\ell}=10$ nodes. The first five points, from left to right, refer to a three layers (linear) neural network. Hence, $\ell=3$ and the size of the intermediate layer is progressively increased, $N_2=20,80,100,500,800$. The total number of trained eigenvalues is $N_2+N_3$, and gets therefore larger as the size of the intermediate layer grows. 
The successive four points of the collections are obtained by setting  $\ell=4$. Here, $N_2=800$ while $N_3$ is varied ($=100,200,400,600$). The training impacts on $N_2+N_3+N_4$ parameters. Finally the last point in each displayed curve is obtained by working with a five layers deep neural network, $\ell=5$. In particular $N_2=800$, $N_3=600$ and $N_4=500$, for a total of $N_2+N_3+N_4+N_5$ tunable parameters. Also in this case,  the spectral algorithm performs better than conventional learning schemes constrained to operate with an identical number of free parameters. Similarly, the distribution of the weights of an equivalent perceptron trained in reciprocal space matches that obtained when operating in the space of the nodes and resting on a considerably larger number of training parameters.  To sum up, eigenvalues are parameters of key importance for neural networks training, way more strategic than any other set of equivalent cardinality in the space of the nodes. As such, they allow for a global approach to the learning, with significant reflexes of fundamental and applied interest. In all cases here considered, the learning can extend to the eigenvectors: an optimized indentation of the eigen-directions contribute to enhance the overall performance of the trained device.}

{We now turn to considering a non-linear architecture. More specifically, we will assume a four layers network with, respectively, $N_1=784, N_2,N_3=120, N_4=10$. The non-linear ReLU filter acts on the third layer of the collection, while the second is a linear processing unit. As in the spirit of the wide network configuration evoked above, we set at testing the performance of the neural network for increasing $N_2$. For every choice of $N_2$, the linear layer can be retracted yielding a three-layered effective non-linear configurations. We recall however that training the network in the enlarged space where the linear unit is present leaves a non trivial imprint in the weights that set the strength of the links in direct space.} 

\begin{figure}
\centering
\includegraphics[scale=0.5]{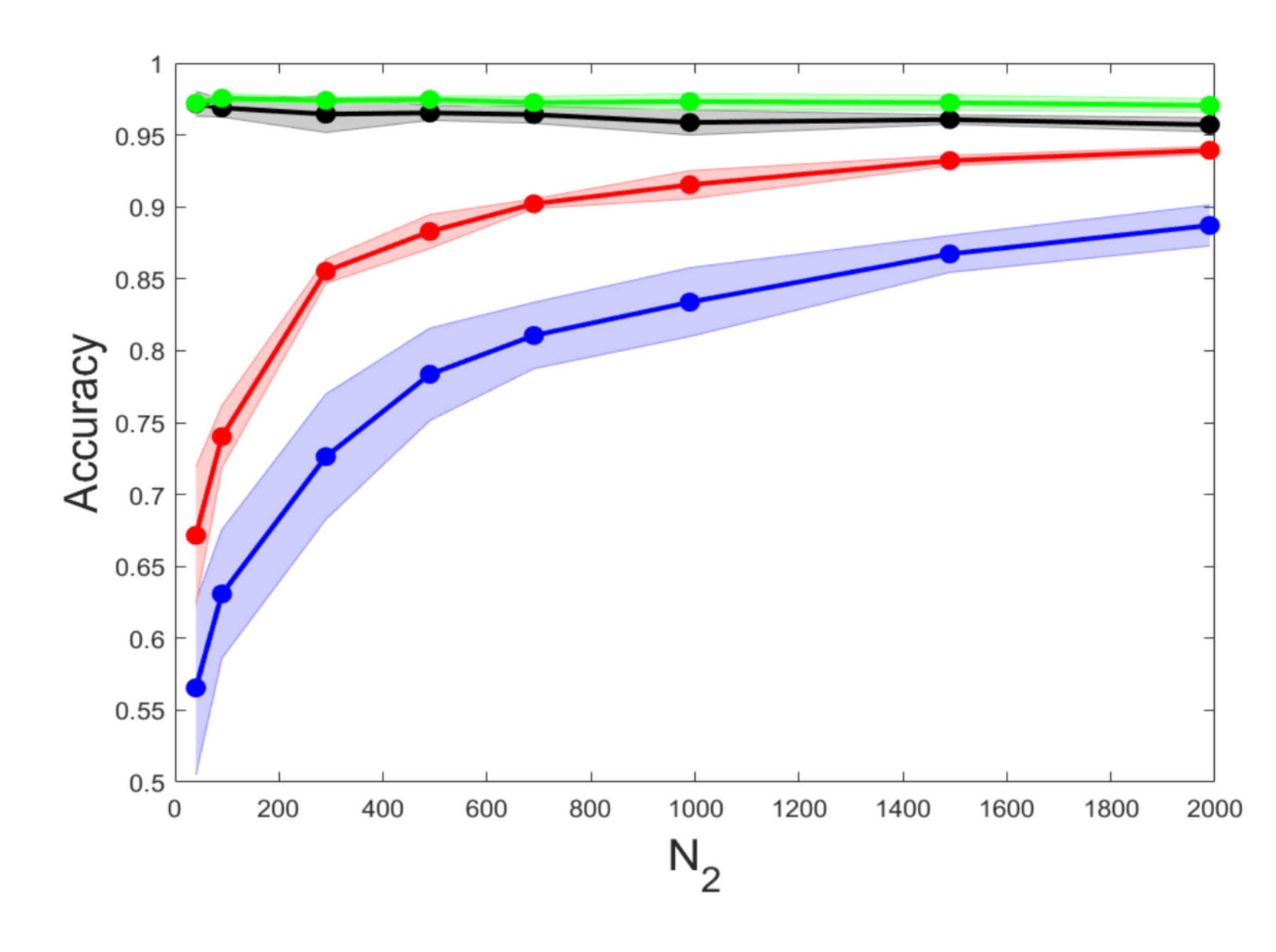} \\ 
\caption{\it  {The accuracy of the non-linear deep neural network is tested. We assume a four layers network with, respectively, $N_1=784, N_2,N_3=120, N_4=10$; $N_2$ is changed so as to enlarge the set of parameters to be trained. The red line refers to the spectral training, with $N_2+N_3+N_4$ adjusted eigenvalues. The blue line stands for a neural network trained in direct space,
the target of the optimization being a subset made of $N_2+N_3+N_4$ weights, randomly selected from the available pool of $N_1 N_2+N_2 N_3+N_3 N_4$ tunable parameters. 
The black line reports the accuracy of the linear neural network when training the full set of $N_1 N_2+N_2 N_3+N_3 N_4$ weights. The green line refer to the spectral learning when eigenvalues and eigenvectors are simultaneously trained.}}
\label{fig7} 
\end{figure}

{In Fig \ref{fig7},  we plot the computed accuracy as a function of $N_2$,  the size of the linear layer. In analogy with the above analysis, the red curve refers to the training restricted to  
$N_2+N_3+N_4$ eigenvalues; the blue profile is obtained when the deep neural network is  trained in direct space by adjusting an identical number of inter-nodes weights. As for the case of a fully linear architecture, by adjusting the eigenvalues yields better classification performances. The black line shows the accuracy of the neural network when the full set of $N_1 N_2+N_2 N_3+N_3 N_4$ is optimized in direct space. The green line refer instead to the spectral learning when the eigenvalues and eigenvectors are trained simultaneously. The accuracies estimated for these two latter settings agree within statistical error, even if the spectral scheme seems more robust to overfitting (the black circles declines slightly when increasing $N_2$, while the collection of green points appears rather stable).}

\section{Conclusions}

Summing up, we have here proposed a novel approach to the training of deep neural networks which is bound to the spectral, hence reciprocal, domain. The {eigenvalues and eigenvectors} of the adjacency matrices that connects consecutive layers via directed feed-forward links are trained, instead of adjusting the weights that bridge each pair of nodes of the collection, as it is customarily done in the framework of conventional ML approaches. 

{The first conclusion of our analysis is that optimizing the eigenvalues, when freezing the eigenvectors, yields performances which are superior to those attained with conventional  methods {\it restricted} to a operate with an identical number of free parameters. It is therefore surmised that eigenvalues are key target parameters for neural networks training, in that they allow for a {\it global} handling of the learning. This is at variance with conventional approaches which seek at modulating the weights of the links among mutually connected nodes. Secondly,  the  spectral learning restricted to the eigenvalues yields a distribution of the weights which resembles quite closely that obtained with conventional algorithms bound to operate in direct space. For this reason, the proposed method could be used in combination with existing ML algorithms for an effective (and computationally advantageous) pre-training of deep neural networks. We have also shown that linear processing units inserted in between consecutive, non-linearly activated layers produce an enlargement of the learning parameters space, with beneficial effects in terms of performance of the trained device. 
Extending the learning so as to optimize the eigenvectors enhances the ability of the network to operate the sought classification.  In the proposed implementation, and to recover a feed-forward architecture in direct space, we have assumed a nested indentation of the eigenvectors. Entangling the eigenvectors referred to successive stacks is the key for a recursive processing of the data, from the input to the output layer.  Employing other non-orthogonal basis could eventually allow to challenge different topologies in direct space and shed novel light on the surprising ability of deep networks to cope with the assigned tasks.} 

{In future perspective, it would interesting to characterize the solutions attained with the spectral method, following the strategy outlined in \cite{feizi2017porcupine}. 
Further, it could be interesting to combine the spectral approach to other existing schemes which have been devised to improve the computational performance of deep neural networks, 
without significant loss in final recognition accuracy \cite{6638949, frankle2020training}.}

\bibliographystyle{unsrt}
\bibliography{bibliography}

\end{document}